\documentclass[11pt]{article}

\usepackage[final]{acl}
\usepackage{times}
\usepackage{latexsym}
\usepackage[T1]{fontenc}
\usepackage[utf8]{inputenc}
\usepackage{microtype}
\usepackage{inconsolata}
\usepackage{graphicx}
\usepackage{subcaption}
\usepackage{booktabs} 
\usepackage{enumitem}
\usepackage{multirow}
\usepackage{amsmath}
\usepackage{amssymb}
\usepackage{mathtools}
\usepackage{amsthm}
\usepackage{algorithm}
\usepackage{algorithmic}

\title{TopicMerge: A Divide and Conquer Approach to Topic Modeling}

\author{
  \textbf{Matthew Raffel}\textsuperscript{1,2},
  \textbf{Lizhong Chen}\textsuperscript{1},
  \textbf{Sai Munikoti}\textsuperscript{2},
  \textbf{Sameera Horawalavithana}\textsuperscript{2} \\
  \textsuperscript{1}Oregon State University \\
  \textsuperscript{2}Pacific Northwest National Laboratory \\
  \texttt{\{raffelm, chenliz\}@oregonstate.edu} \\
  \texttt{\{sai.munikoti, sameera.horawalavithana\}@pnnl.gov}
}

\begin{document}
\maketitle
\begin{abstract}
Recent advancements in topic modeling have split into two main paths: efficient embedding-based methods that offer scalability but are often outperformed on standard clustering metrics, and Large Language Model (LLM) approaches that achieve state-of-the-art clustering performance but at the cost of prohibitive computational expense and poor scalability. This paper introduces \textit{TopicMerge}, a novel multi-stage pipeline that bridges these approaches to achieve both state-of-the-art clustering performance and computational efficiency. TopicMerge first employs an efficient clustering pipeline to deliberately over-segment a document corpus into a large set of fine-grained, semantically pure micro-clusters, which are then given initial semantic labels. It then introduces a novel, iterative merging algorithm where an LLM, guided by local neighborhood context, arbitrates on consolidating candidate pairs. These approved merges are then selectively executed and semantically relabeled by the LLM, hierarchically building broader, more coherent topics. Extensive experiments on the Bills and Wiki datasets show that TopicMerge achieves state-of-the-art clustering performance while reducing required LLM calls by over 92\% compared to document-level methods, offering a practical and scalable solution.
\end{abstract}

\section{Introduction}
The ability to automatically discover thematic structures from large text corpora is a fundamental task in natural language processing (NLP). Topic modeling, the discipline dedicated to this task, has evolved significantly. Early methods such as Latent Dirichlet Allocation (LDA) \cite{blei2003latent} and Non-negative Matrix Factorization (NMF) \cite{lee1999learning, fevotte2011algorithms} established a strong foundation by treating documents as mixtures of topics. However, their reliance on a bag-of-words representation limited their ability to capture deeper semantic context. The advent of pre-trained language models (PLMs) led to a new paradigm of embedding-based topic models, such as BERTopic \cite{grootendorst2022bertopic}, which cluster documents in a rich semantic space to produce more coherent topics. Most recently, the powerful reasoning capabilities of Large Language Models (LLMs) have been leveraged directly for topic discovery, with models like TopicGPT \cite{pham2024topicgpt} and LiSA \cite{liu-etal-2025-llm-guided} setting a new state-of-the-art, achieving superior performance on standard clustering metrics.

Although LLM-based topic models offer a leap in clustering performance, they also introduce two fundamental challenges. First, they suffer from high computational costs and poor scalability. Each LLM inference consumes substantial resources, and this cost becomes prohibitive in models requiring per-document interactions, as the total expense scales with dataset size. Second, and more fundamentally, their processing is inherently local. Constrained by a fixed context window, an LLM evaluating a single document or a small batch lacks a global view of the corpus's thematic structure. It cannot natively perceive the relationships between all documents at once, a capability essential for comprehensive topic discovery. This creates a critical research gap: how to leverage the powerful local reasoning of LLMs without inheriting their prohibitive costs and their lack of a global perspective.

To bridge this gap, we introduce \textit{TopicMerge}\footnote{Our code is publicly available at \url{https://github.com/pnnl/PermitAI/tree/master/TopicMerge}.}, a novel three-stage hybrid pipeline that combines the global structural awareness of clustering with the local semantic reasoning of an LLM. The process begins with an initial clustering stage, where an efficient clustering algorithm deliberately over-segments the corpus into fine-grained, pure micro-clusters, establishing a global structure. Following this, an initial labeling stage uses an LLM to assign a concise label and description to each micro-cluster. The core of our contribution is the third stage consisting of a hierarchical merging process where an LLM performs two key roles within a multi-phase algorithm. First, it acts as a context-aware semantic arbiter, evaluating individual candidate pairs to decide whether they should be consolidated. From this pool of approved candidates, the algorithm then prioritizes the most cohesive pairings for execution. It then acts as a summarizer, synthesizing the contents of the original clusters to generate a new, more abstract label for each emergent topic.  This hierarchical approach, leveraging the LLM for both high-level arbitration and summarization, enables nuanced semantic decisions while efficiently operating at the cluster level.

To validate the effectiveness and efficiency of TopicMerge, we conduct extensive experiments on two widely-used benchmark datasets, Bills and Wiki \cite{hoyle2022neural}. We benchmark our model against a comprehensive suite of baselines spanning traditional (LDA) \cite{blei2003latent}, embedding-based (BERTopic) \cite{grootendorst2022bertopic}, and state-of-the-art LLM-based (TopicGPT, LiSA) paradigms \cite{pham2024topicgpt, liu-etal-2025-llm-guided}. Our evaluation focuses on two key dimensions: clustering quality, measured by standard metrics including P1 \cite{amigo2009comparison}, ARI \cite{vinh2009information} and NMI \cite{strehl2002cluster}, and computational efficiency, measured by the total number of LLM prompts required. Furthermore, a detailed ablation study dissects our model's core components, empirically validating the strategic importance of contextual clusters and beginning merging with fine-grained, pure clusters.

Our main contributions are as follows:
\begin{enumerate}[topsep=0pt]
    \item We propose TopicMerge, a novel three-stage pipeline that effectively integrates initial over-clustering with subsequent LLM-guided hierarchical consolidation.
    \item We introduce a context-aware LLM merging algorithm that operates efficiently at the cluster level, enabling nuanced semantic decisions without the cost of per-document analysis.
    \item We demonstrate through extensive experiments that TopicMerge achieves state-of-the-art clustering performance on widely-used benchmark datasets.
    \item We show that our method provides a practical path to leveraging LLMs by reducing the number of required prompts by over 92\% compared to alternative document-level LLM-based approaches, drastically lowering overall computational cost and enabling high-quality topic modeling to scale.
\end{enumerate}

\section{Related Works}
\label{sec:related_works}

\subsection{Traditional Topic Modeling}
Foundational topic modeling was dominated by two main approaches: probabilistic generative models and matrix factorization. The former is exemplified by the seminal LDA \cite{blei2003latent}, a Bayesian model that represents documents as mixtures of latent topics and generates words under a bag-of-words assumption. The latter is represented by NMF \cite{lee1999learning, fevotte2011algorithms}, which offers a linear-algebraic alternative by decomposing a document-term matrix. Despite their different mathematical foundations, both LDA and NMF are fundamentally limited by the bag-of-words assumption. As a result, they are unable to capture word order and semantic context, which often leads them to conflate distinct word meanings, resulting in topics that lack deep semantic coherence.

\subsection{Embedding-based Topic Modeling}
The advent of PLMs such as BERT \cite{devlin2019bert} and RoBERTa \cite{liu2019roberta} enabled a new paradigm of topic modeling centered on clustering contextualized document embeddings. The general approach involves encoding documents into dense vectors, often using architectures specialized for sentence similarity, such as Sentence-BERT \cite{reimers2019sentence}. This pipeline is exemplified by influential models such as Top2Vec \cite{angelov2020top2vec} and the highly modular BERTopic \cite{grootendorst2022bertopic}, which typically employ dimensionality reduction (e.g., UMAP \cite{mcinnes2018umap}) and density-based clustering (e.g., HDBSCAN \cite{malzer2020hybrid, mcinnes2017accelerated}) to identify topics from the resulting embeddings. While subsequent work, such as FASTopic \cite{wu2024fastopic}, has focused on optimizing the speed and stability of this process, a shared limitation persists. The reliance on purely geometric clustering, while producing coherent groups, often yields generic topics. This limitation creates an opening for LLM-based approaches, which can leverage their contextual understanding to both guide the formation of clusters and provide inherently interpretable representations.

\subsection{Topic Modeling with Large Language Models}
The most recent evolution in topic modeling leverages the advanced reasoning and language understanding capabilities of LLMs such as GPT-4 and Gemini 2.5 \cite{achiam2023gpt, comanici2025gemini}. These methods move beyond clustering embeddings and instead interact with an LLM directly to perform topic discovery.

One prominent approach, prompt-based topic modeling, delegates key interpretation tasks to an LLM, with methods like TopicGPT \cite{pham2024topicgpt} framing the entire process as a conversation for discovery and summarization, while others like PromptTopic \cite{wang2023prompting} focus on generating coherent labels for existing document clusters. A secondary growing body of work employs LLMs as specialized components within a larger, often traditional, pipeline aiming to combine the strengths of different methods.  For instance, by using an LLM to actively guide the core clustering algorithm as in LISA \cite{liu-etal-2025-llm-guided} or k-LLMmeans \cite{diaz2025k}, post-processing and refinement as in EDTM \cite{dhanania2024interactive}, or determining an optimal clustering granularity of an hierarchical clustering algorithm by posing pairwise questions to an LLM as in ClusterLLM \cite{zhang-etal-2023-clusterllm}, state-of-the-art results are achievable.

However, a unifying challenge for these LLM-centric approaches lies in how they interact with the LLM, which introduces a fundamental trade-off between scope and efficiency. On one hand, to manage costs, an LLM's judgments are made with an inherently local and fragmented view (e.g., one document), reasoning about isolated data without access to the broader thematic context of the entire corpus \cite{zhang-etal-2023-clusterllm, pham2024topicgpt}. On the other hand, attempting to provide a more global view by increasing the number or complexity of queries can lead to significant scalability hurdles and prohibitive computational costs \cite{liu-etal-2025-llm-guided, dhanania2024interactive, pham2024topicgpt, wang2023prompting}. This forces existing methods to either sacrifice global awareness for efficiency or face impractical costs. Our work is positioned to directly address this challenge by combining the global awareness of efficient clustering with the targeted, local semantic reasoning of an LLM.

\section{Methods}

\begin{figure*}[t!] 
    \centering
    \includegraphics[width=0.95\textwidth]{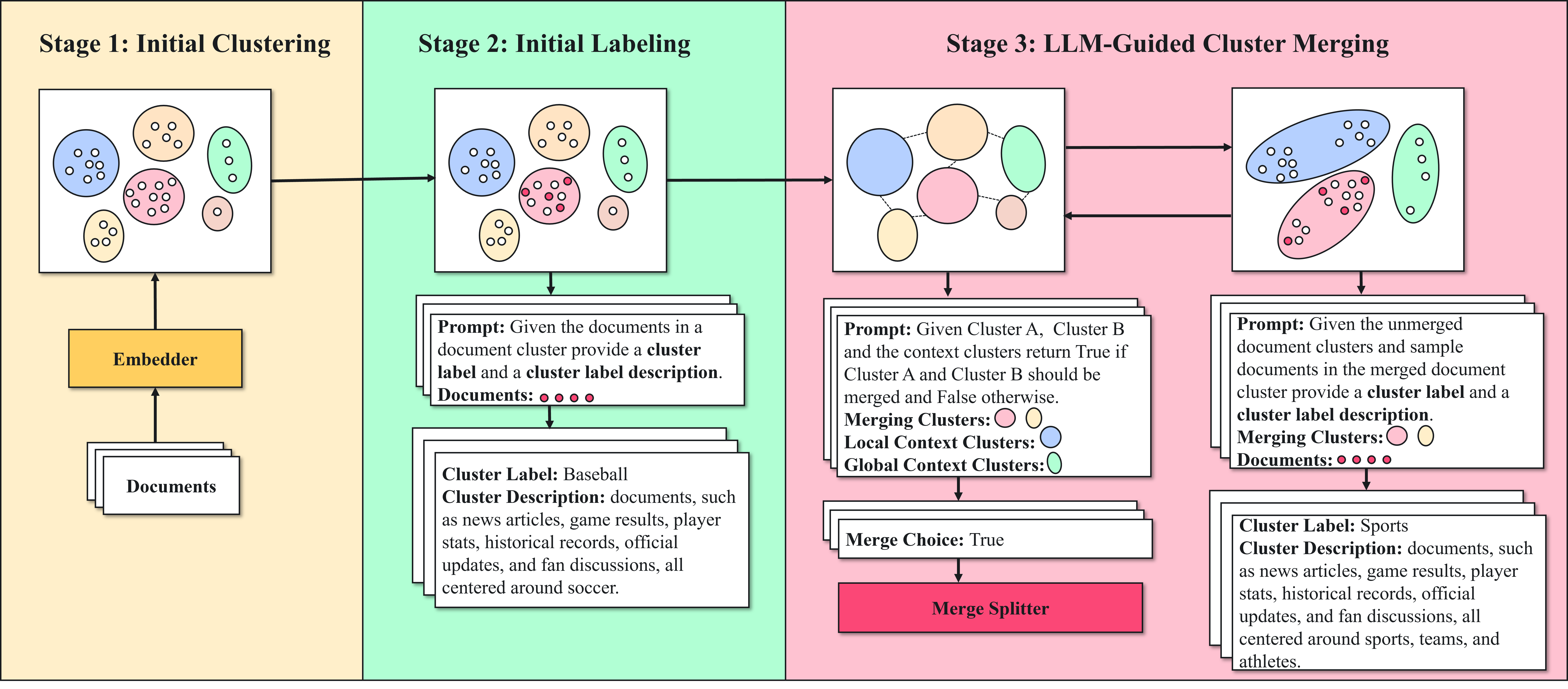} 
    \caption{An overview of TopicMerge. The process begins with (1) initial clustering stage, where document embeddings are deliberately over-segmented into numerous, fine-grained micro-clusters. These are then given preliminary semantic representations in the (2) initial labeling stage. Then in the (3) a LLM-guided cluster merging stage, an LLM acts as a context-aware semantic arbiter to hierarchically consolidate the micro-clusters.}
    \label{fig:topicmerge}
\end{figure*}
In this section, we present TopicMerge, a multi-stage pipeline for topic modeling that integrates traditional clustering techniques with LLMs for intelligent merging and labeling. The core of our approach is to first use an efficient clustering algorithm to establish a global, structural understanding of the corpus and then leverage an LLM for targeted, high-level semantic judgments. Critical to this design is the recognition that the quality of the LLM's judgment hinges on the clarity of its input. By deliberately creating fine-grained, semantically pure micro-clusters, we provide the LLM with unambiguous concepts to evaluate, which significantly improves the accuracy of its merging decisions. This hierarchical approach not only improves topic modeling but also significantly reduces the number of required LLM calls by operating at the cluster level, making the process highly scalable.

As visualized in Figure~\ref{fig:topicmerge}, the pipeline consists of three main stages: (1) an initial clustering stage, where we generate document embeddings and cluster them to form the set of pure micro-clusters; (2) an initial labeling stage, which produces a label and description for each cluster; and (3) an LLM-guided cluster merging stage that iteratively merges these clusters in a hierarchical fashion. We will now explain each stage in more detail.

\subsection{Initial Clustering Stage}
The first stage involves generating an initial set of clusters from a collection of raw input text data. This process comprises three sub-steps: embedding creation, dimensionality reduction, and clustering.  The main objective of the initial clustering stage is to create a set of pure clusters.  As such, all the sub-stages are designed to produce a large number of these pure clusters.  In doing so, it will allow the following labeling and merging stages to be less influenced by noisy clusters.

Input text documents are first converted into an embedding representation using a pre-trained embedding model, building on the methodology of BertTopic \cite{grootendorst2022bertopic}. The resulting embeddings are then normalized to unit length. While the full-dimensional embeddings are retained for semantic calculations in later stages, a targeted dimensionality reduction is applied specifically to prepare the data for the initial clustering step. This pre-processing improves computational efficiency, reduces semantic noise, and enhances the stability of the clustering task \citep{amate2026assessing}. We employ Pairwise Controlled Manifold Approximation (PaCMAP) \citep{JMLR:v22:20-1061} for this purpose, as its design balances the preservation of both local and global structure, which is crucial for identifying robust, fine-grained micro-clusters.

Following dimensionality reduction, we perform clustering to generate the initial set of micro-clusters. This stage can be performed by any algorithm capable of producing a specified number of clusters. For our experiments, we use Agglomerative Clustering to generate these compact clusters with cosine distance and a complete linkage criterion. To determine the initial cluster count, we employ a specific strategy for over-segmentation. First, we use the Silhouette score not to find an optimal $k$, but to establish a lower bound for a plausible number of topics \cite{rousseeuw1987silhouettes}. We then intentionally set the actual cluster count to a value significantly exceeding this baseline. This deliberate over-segmentation yields a large set of fine-grained, semantically pure micro-clusters, providing an ideal and unambiguous foundation for the subsequent LLM-guided cluster merging stage.

\subsection{Initial Labeling Stage}
Following the initial generation of clusters, we label each cluster with a concise, human-readable label and a comprehensive description. These semantic representations would serve as the primary input for the subsequent LLM-guided merging phase, allowing it to make decisions based on semantic understanding. To generate these representations, a dedicated labeling module first selects a subset of representative documents from each cluster. During the selection, we use the Farthest Point Sampling (FPS) on the document embeddings. Unlike sampling documents near the cluster centroid, which might over-represent the core theme, FPS greedily selects a fixed number (e.g., 32) of diverse examples that are maximally distant from one another. This provides a comprehensive overview of the cluster's semantic space. 

These sampled texts are then formatted into a structured prompt for an LLM, which is instructed to synthesize the common themes across the diverse examples and produce both a concise label (e.g., ``Medical Bills'') and a detailed description (e.g., ``Documents related to medical services, billing, and insurance claims.''). While the FPS sampling strategy captures broad coverage of the cluster's content, the prompt's instruction synthesizes specificity in the final label.  Furthermore, to ensure the breadth of the topics are limited we also provide some example topic labels from the dataset. This staged-approach prevents the generation of labels that are either too generic to be useful or too narrow to describe a significant fraction of the cluster's documents.

\subsection{LLM-Guided Cluster Merging Stage}

In this stage, we transform the initial set of fine-grained, pure clusters into a final set of broader, semantically coherent topics. This is an iterative consolidation process where an LLM serves a dual role: first as a semantic arbiter making nuanced decisions about which clusters to merge, and then as a label synthesizer to generate a new, more abstract topic for the merged cluster. The entire procedure, detailed in Algorithm \ref{alg:llm_merging}, is designed to balance semantic accuracy with computational efficiency. We will now walk through the four key phases of each iteration: (1) efficient candidate generation, (2) context-aware LLM arbitration, (3) purity-based ranking and independent set selection, and (4) merge execution with relabeling.

\begin{algorithm}[h!]
    \small
   \caption{LLM-Guided Hierarchical Cluster Merging}
   \label{alg:llm_merging}
\begin{algorithmic}[1]
   \REQUIRE Initial clusters $\mathcal{C}_{init}=\{C_1, ..., C_M\}$, document embeddings $E=\{e_1, ..., e_N\}$, LLM Oracle $\mathcal{L}$, Relabeling Oracle $\mathcal{R}$, k-NN parameter $k$, and example topics $T$.
   \STATE Initialize Union-Find structure $U$ with $\mathcal{C}_{init}$.
   \STATE Initialize LLM decision cache $\mathcal{K} \leftarrow \emptyset$.
   \STATE $merges\_made \leftarrow \text{True}$
   \WHILE{$merges\_made$ is True}
      \STATE $merges\_made \leftarrow \text{False}$
      \STATE $\mathcal{C}_{curr} \leftarrow$ set of all current root clusters from $U$
      \STATE $G \gets \text{construct\_augmented\_mst}(\mathcal{C}_{curr}, E, k)$
      \STATE $\mathcal{M}_{approved} \leftarrow \emptyset$
      \FOR{each pair $(C_A, C_B) \in G$}
        \STATE $\mathcal{C}_{local}, \mathcal{C}_{global} \leftarrow f_{context}(C_A, C_B, E)$
        \STATE $P \leftarrow \text{create\_prompt}(C_A, C_B, \mathcal{C}_{local}, \mathcal{C}_{global}, T)$
        \IF{$(C_A, C_B, \mathcal{C}_{local}) \in \mathcal{K}$}
            \STATE $should\_merge \leftarrow \mathcal{K}[(C_A, C_B, \mathcal{C}_{local})]$
        \ELSE
            \STATE $should\_merge \leftarrow \mathcal{L}(P) \wedge \mathcal{L}(P)$ 
            \STATE $\mathcal{K}[(C_A, C_B, \mathcal{C}_{local})] \leftarrow should\_merge$
        \ENDIF
        \IF{$should\_merge$ is True}
            \STATE $s \leftarrow \text{purity\_score}(C_A, C_B, E)$
            \STATE Add $(C_A, C_B, s)$ to $\mathcal{M}_{approved}$
        \ENDIF
      \ENDFOR
      \IF{$\mathcal{M}_{approved}$ is not empty}
          \STATE Sort $\mathcal{M}_{approved}$ by score $s$ in descending order
          \STATE $\mathcal{M}_{exec} \leftarrow \text{greedy\_independent\_set}(\mathcal{M}_{approved})$
          \STATE $merges\_made \leftarrow \text{True}$
          \FOR{each pair $(C_A, C_B) \in \mathcal{M}_{exec}$}
            \STATE $C_{new} \leftarrow U.union(C_A, C_B)$
            \STATE $\text{label}(C_{new}) \leftarrow \mathcal{R}(C_A, C_B, E, T)$
          \ENDFOR
      \ENDIF
   \ENDWHILE
   \STATE \textbf{return} $\mathcal{C}_{final}$
\end{algorithmic}
\end{algorithm}

\subsubsection{Efficient Candidate Generation}
We efficiently identify potential merge candidates in each iteration to circumvent a computationally expensive quadratic comparison of all cluster pairs. We do this by generating a targeted set of candidates by constructing an augmented Reciprocal Nearest Neighbor (RNN) graph, $G$ (line 7). In this graph, an edge initially connects two clusters only if they are mutual k-nearest neighbors (based on centroid cosine distance), which identifies high-confidence pairings. However, as this can produce a disconnected graph, we augment it with minimum-cost ``bridge'' edges to guarantee connectivity between all components. The resulting graph balances sparsity with full coverage, drastically reducing the search space and focusing the LLM's expensive reasoning capabilities only on the most plausible merge candidates.

\subsubsection{Context-Aware LLM Arbitration}
For each candidate pair $(C_A, C_B)$ an LLM is queried to arbitrate the merge decision. A simple pairwise comparison is insufficient, as the semantic relationship between two topics can be ambiguous in isolation. For instance, while ``Soccer'' and ``Basketball'' might not seem mergeable alone, the presence of a neighboring ``Baseball'' cluster provides the necessary context for the LLM to infer a broader, unifying ``Sports'' theme. 

Capitalizing on this insight, we provide the LLM with a rich semantic context beyond just the candidate pair. The prompt includes a local context ($\mathcal{C}_{local}$) of the nearest neighboring clusters, a global context ($\mathcal{C}_{global}$) of more distant representative clusters, and a set of high-level topic examples from the domain to calibrate its sense of appropriate granularity. The LLM is prompted only with cluster labels and descriptions, not raw text, to ensure it reasons about overarching themes rather than minor textual differences. To improve reliability and efficiency, we employ a self-consistency mechanism where merges require two-fold agreement from the LLM, and all decisions are cached to prevent redundant queries (lines 15-16).

\subsubsection{Purity Ranking and Set Selection}
After the LLM has evaluated all candidate pairs, the approved merges are not executed indiscriminately. Instead, each approved pair is assigned a purity score that measures how geometrically indistinguishable the two clusters are in the embedding space (line 19). Specifically, we compute a local silhouette-based score on the subset of points belonging to the two clusters: a high score indicates that the two clusters are already well-mixed and difficult to separate, suggesting a natural merge. The approved merges are then sorted by this purity score in descending order (line 24). From this ranked list, we greedily select a maximal independent set of non-conflicting merges so that no cluster participates in more than one merge within a single iteration (line 25). This prevents cascading merges that could lead to unstable behavior and ensures that the decisions made within a single pass are independent of one another.

\subsubsection{Merge Execution and Relabeling}
The selected, non-conflicting merges are then executed in a single pass using a highly efficient Union-Find data structure, with a union operation grouping each approved pair under a single parent identifier (lines 27-29). Immediately following this structural merge, a semantic update occurs where each newly formed super-cluster is relabeled by the LLM oracle, $\mathcal{R}$ (line 29). To generate a new label that captures the broader theme, the oracle is provided with the source clusters' labels and descriptions along with FPS-sampled datapoints from the combined set and some example topic labels from the dataset. This relabeling ensures that cluster representations remain semantically accurate as they grow.

Once all merges and relabeling are complete, a new, smaller set of clusters forms the input for the next iteration. This entire cycle consisting of candidate generation, context-aware LLM querying, purity-based selection, and merge execution with relabeling repeats. The process naturally converges when a full iteration completes with no new merges being approved by the LLM. This stable state indicates that the topic structure has been consolidated as far as is semantically justified by the LLM, and the final set of clusters is returned.

\section{Experiments}
TopicMerge offers state-of-the-art clustering capabilities while requiring fewer LLM prompts than alternative approaches.  We demonstrate this with extensive experiments and an ablation study.  Firstly, in Section \ref{sec:experimental_setup} we provide our experimental setup. Then, in Section \ref{sec:cluster_results} and \ref{sec:llm_prompt_results}, we demonstrate TopicMerge's state-of-the-art clustering capabilities with minimal LLM prompts.  Finally, in Section \ref{sec:ablations} we provide extensive ablations.

\subsection{Experimental Setup}
\label{sec:experimental_setup}
\subsubsection{Dataset}
We evaluate TopicMerge on \textit{Bills} \cite{hoyle2022neural} and \textit{Wiki} \cite{merity2016pointer, hoyle2022neural}. Bills contains 8,024 summaries of bills from the 110th to 114th U.S. Congresses, while the \textit{Wiki} dataset consists of 15,242 Wikipedia articles. Each corpora has been manually annotated, with Bills and Wiki assigned 21 and 15 high-level topic labels, respectively. 

\subsubsection{Evaluation Metrics}
Following prior work \cite{hoyle2022neural, wu2024fastopic, liu-etal-2025-llm-guided, pham2024topicgpt}, we measure the alignment between generated document clusters and the ground-truth labels using the purity (P) \cite{zhao2001criterion}, the harmonic mean of purity and inverse purity (P1) \cite{amigo2009comparison}, Adjusted Rand Index (ARI) \cite{vinh2009information}, and Normalized Mutual Information (NMI) \cite{strehl2002cluster}.  By choosing these four metrics, we offer complementary perspectives of the clusters.  
While P focuses on the frequency of the most common category in each cluster, P1 provides a more robust metric that also accounts for inverse purity, rewarding grouping items together without penalizing mixing different categories.  ARI measures the pairwise agreement between sets while adjusting for chance \cite{vinh2009information}.  NMI measures the information shared between the ground-truth and predicted clusters while accounting for class imbalance \cite{strehl2002cluster}. We provide equations for these metrics in Appendix \ref{app:eval_metrics}.

\subsubsection{Baselines}
We compare TopicMerge against a comprehensive set of prior models (9 in total) that span traditional, embedding-based, and LLM-based paradigms. The traditional models include the LDA \cite{blei2003latent} and NMF \cite{lee1999learning}. For neural and embedding-based approaches, we select Contextualized Topic Models (CTM) \cite{bianchi-etal-2021-cross}, BERTopic \cite{grootendorst2022bertopic}, and FASTopic \cite{wu2024fastopic}. Finally, we compare against state-of-the-art LLM-based models: EDTM \cite{dhanania2024interactive}, PromptTopic \cite{wang2023prompting}, TopicGPT \cite{pham2024topicgpt}, and LiSA \cite{liu-etal-2025-llm-guided}. 

\subsubsection{Hyperparameters}
\label{sec:hyper}
For creating embeddings of text documents, we generate 256-dimensional Gemini clustering embeddings \cite{lee2025gemini}.  We reduce the dimensionality using PaCMAP \cite{JMLR:v22:20-1061} to 64 dimensions.  Our PaCMAP hyperparameters consist of a mid-near and further pair ratio of 0.5 and 2, respectively.  The number of neighbor points is set to $10 + 15 * (\log_{10}(N) - 4)$ where $N$ is the dataset size.\footnote{The default setting from the PaCMAP library: \url{https://github.com/YingfanWang/PaCMAP}} We then applied Agglomerative Clustering to form an initial set of 80 clusters, using a cosine distance metric and complete linkage.  For the labeling and merging stages, our primary experiments, prompt counts, and ablations use Gemini 2.5 Flash Lite \cite{comanici2025gemini} with temperature set to 0. In addition, we evaluate the newer Gemini 3.1 Flash Lite in our main results (Table~\ref{tab:results_simple}) to examine model generalization and reasoning improvements.  For all our prompting, we use the DSPy framework \cite{khattab2024dspy}.  When labeling, we use a batch size of 32 samples per cluster. During the merging stage, we create our RNN graph with 5 neighbors.  For each LLM prompt we provide 3 and 5 local and global context clusters, respectively.  We also provide 2-3 example topic labels. We provide the prompts for our labeling and merging stages in Appendix \ref{app:prompts}.

\subsection{Clustering Results}
\label{sec:cluster_results}
As presented in Table~\ref{tab:results_simple}, TopicMerge achieves state-of-the-art or highly competitive performance across both benchmark datasets. On the Wiki dataset, TopicMerge is the top-performing model, establishing a new state-of-the-art with an ARI of 0.71, an NMI of up to 0.74, and a P1 of 0.73. This represents a significant improvement over all baselines, including strong LLM-based methods like LiSA.
\begin{table}[ht]
\centering
\small
\setlength{\tabcolsep}{3pt}
\caption{Experimental results comparing TopicMerge against baselines. The best results are denoted in \textbf{bold} and the second-best results are marked by \underline{underline}.}
\label{tab:results_simple}
\begin{tabular}{@{}l ccc ccc@{}}
\toprule
& \multicolumn{3}{c}{\textbf{Bills}} & \multicolumn{3}{c}{\textbf{Wiki}} \\
\cmidrule(lr){2-4} \cmidrule(lr){5-7}
& \textbf{P1} & \textbf{ARI} & \textbf{NMI} & \textbf{P1} & \textbf{ARI} & \textbf{NMI} \\
\midrule
LDA & 0.52 & 0.31 & 0.42 & 0.72 & 0.63 & 0.71 \\
NMF & 0.24 & 0.04 & 0.15 & 0.58 & 0.38 & 0.55 \\
CTM & 0.37 & 0.19 & 0.32 & 0.61 & 0.46 & 0.62 \\
FASTopic & 0.40 & 0.30 & 0.33 & 0.65 & 0.56 & 0.60 \\
EDTM & \textbf{0.58} & — & — & 0.65 & — & — \\
BERTopic & 0.41 & 0.18 & 0.37 & 0.62 & 0.50 & 0.59 \\
PromptTopic & 0.36 & 0.17 & 0.28 & 0.62 & 0.46 & 0.59 \\
TopicGPT  & \underline{0.57} & \underline{0.40} & 0.49 & \textbf{0.74} & 0.60 & 0.70 \\
LiSA & \underline{0.57} & \underline{0.40} & 0.51 & \textbf{0.74} & \underline{0.66} & 0.71 \\
\midrule
\multicolumn{7}{@{}l}{\textbf{TopicMerge (Ours)}} \\
\quad \textit{w/ 2.5 Flash-Lite} & \underline{0.57} & \textbf{0.42} & \textbf{0.53} & 0.72 & \textbf{0.71} & \underline{0.73}\\
\quad \textit{w/ 3.1 Flash-Lite} & \underline{0.57} & \textbf{0.42} & \underline{0.52} & \underline{0.73} & \textbf{0.71} & \textbf{0.74}\\
\bottomrule
\end{tabular}
\end{table}

On the more complex Bills dataset, TopicMerge demonstrates robust performance, achieving the best NMI (0.53) and ARI (0.42) across all approaches, with a P1 score of 0.57. Comparing the two LLM backbones, while Gemini 2.5 Flash Lite already outperforms prior baselines, Gemini 3.1 Flash Lite achieves further gains (e.g., reaching 0.74 NMI on Wiki). Importantly, we found that Gemini 3.1 Flash Lite was noticeably more consistent in arbitrating merge decisions and synthesizing appropriate multi-level abstractions, leading to a higher-quality topic hierarchy overall. We provide the complete final topic labels produced by our default Gemini 2.5 Flash Lite pipeline in Appendix~\ref{app:topics}.

\subsection{LLM Prompt Scaling and Efficiency}
\label{sec:llm_prompt_results}

A key advantage of TopicMerge is its computational efficiency and scalability, particularly in its use of the LLM. Table \ref{tab:multilevel_header} compares the number of required LLM calls for TopicMerge against other LLM-based approaches. Approaches like LiSA, which may interact with the LLM on a per-document or per-batch basis, require a number of prompts that scales with the size of the dataset, making them prohibitively expensive for large corpora (over 8,000 and 15,000 prompts for Bills and Wiki, respectively).
\begin{table}[h!]
    \centering
    \small
    \caption{A reporting of the number of prompts required for an LLM to classify each dataset.  The number of prompts reported for PromptTopic, TopicGPT, and LiSA are lower-bounds based on their respective algorithms.}
    \label{tab:multilevel_header}
    \setlength{\tabcolsep}{3pt}
    \begin{tabular}{l cccc} 
        \toprule

        \textbf{} & \multicolumn{4}{c}{\textbf{Models}} \\ 
        \cmidrule(lr){2-5} 
        
        & PromptTopic & TopicGPT & LiSA & TopicMerge \\
        \midrule
        
        Wiki   & $15,242+$ & $15,242+$ & $15,242+$ & 562 \\
        Bills & $8,024+$ & $8,024+$ & $8,024+$ & 645 \\
        
        \bottomrule
    \end{tabular}
\end{table}

In contrast, TopicMerge's LLM interaction is limited to the cluster-level merging and labeling stages. By first creating initial pure clusters and only querying the LLM to make high-level decisions about merging them, our method reduces the number of required LLM calls. For the Bills and Wiki datasets, TopicMerge (using Gemini 2.5 Flash Lite) required only 645 and 562 prompts, respectively, resulting in a reduction of over 92\% compared to document-level LLM approaches. This demonstrates our method provides a practical and scalable path to leveraging the power of LLMs for topic modeling on real-world datasets.

\subsection{Ablations}

\label{sec:ablations}
We conduct ablations of TopicMerge analyzing the importance of having initial pure cluster from the initial clustering stage and the influence of context clusters in the LLM-guided cluster merging stage.

\subsubsection{Context Cluster Ablation}

The results in Table \ref{tab:context} highlight the importance of providing semantic context. On the Wiki dataset, removing context clusters causes a significant performance drop (e.g., ARI from 0.71 to 0.63), showing it is critical for disambiguating the corpus's nuanced and overlapping topics. This trend holds on the Bills dataset, further validating our hypothesis that contextual awareness is a key factor in enabling the LLM to make robust, globally-informed merging decisions.

\begin{table}[h]
\centering
\small
\caption{Ablation results of our TopicMerge with and without context clusters.}
\label{tab:context}
\begin{tabular}{@{}l ccc ccc@{}}
\toprule
& \multicolumn{3}{c}{\textbf{Bills}} & \multicolumn{3}{c}{\textbf{Wiki}} \\
\cmidrule(lr){2-4} \cmidrule(lr){5-7}
& \textbf{P1} & \textbf{ARI} & \textbf{NMI} & \textbf{P1} & \textbf{ARI} & \textbf{NMI} \\
\midrule
TopicMerge & 0.57 & 0.42 & 0.53 & 0.72 & 0.71 & 0.73\\
 \hspace{10pt} - Context & 0.56 & 0.41 & 0.53 & 0.70 & 0.63 & 0.70\\
\bottomrule
\end{tabular}
\end{table}

\subsubsection{Cluster Count and Purity Ablation}

Table \ref{tab:cluster_count} analyzes the impact of initial cluster granularity on final topic quality. In this experiment, we vary the initial cluster count (CC) from 80 down to 30. As a higher cluster count produce smaller and more semantically pure clusters, this allows us to examine how the purity (P) of the starting clusters affects the final metrics (P1, ARI, NMI) after the LLM-guided merging stage is complete.

\begin{table}[h]
\centering
\small
\caption{A clustering quality comparison with the merging stage across different cluster counts from the initial clustering. The \textbf{CC} and \textbf{P} columns show the initial cluster count and corresponding purity without merging.}
\label{tab:cluster_count}
\setlength{\tabcolsep}{3pt}
\begin{tabular}{@{}cc|ccc|cc|ccc@{}}
\toprule
\multicolumn{5}{c|}{\textbf{Bills}} & \multicolumn{5}{c}{\textbf{Wiki}} \\
\cmidrule(lr){1-5} \cmidrule(lr){6-10}
\textbf{CC} & \textbf{P} & \textbf{P1} & \textbf{ARI} & \textbf{NMI} & \textbf{CC} & \textbf{P} & \textbf{P1} & \textbf{ARI} & \textbf{NMI} \\
\midrule
80 & 0.72 & 0.57 & 0.42 & 0.53 & 80 & 0.88 & 0.72 & 0.71 & 0.73 \\
60 & 0.71 & 0.56 & 0.43 & 0.52 & 60 & 0.87 & 0.73 & 0.71 & 0.72 \\
40 & 0.68 & 0.58 & 0.41 & 0.51 & 40 & 0.82 & 0.68 & 0.57 & 0.69 \\
30 & 0.62 & 0.51 & 0.37 & 0.47 & 30 & 0.79 & 0.62 & 0.50 & 0.67 \\
\bottomrule
\end{tabular}
\end{table}

\begin{figure}[t!] 
    \centering

    \begin{subfigure}[h]{0.23\textwidth}
        \includegraphics[width=\linewidth]{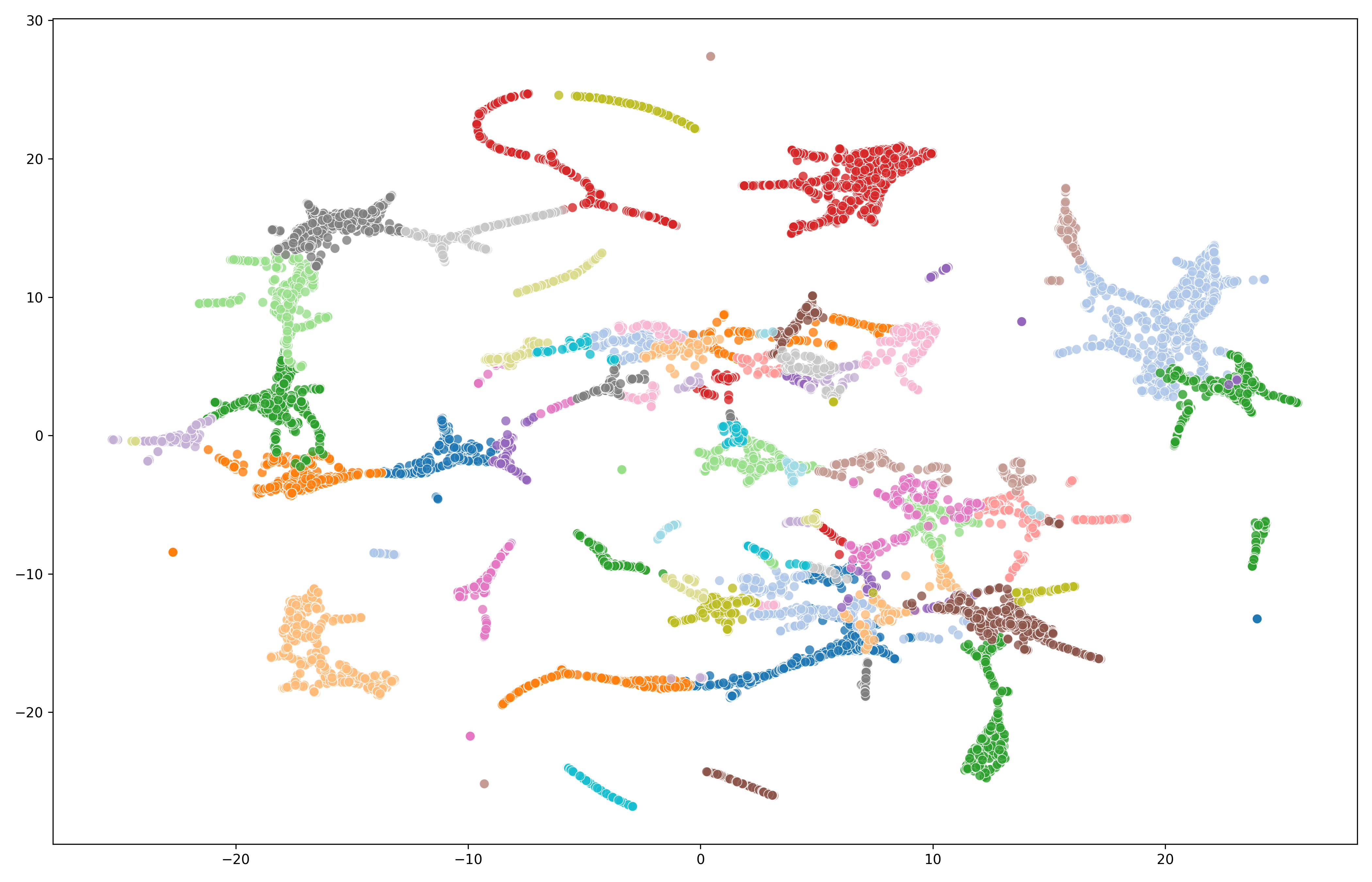}
        \caption{Embeddings for the initial cluster assignments.}
        \label{fig:cluster_visuals1}
    \end{subfigure}
    \hfill
    \begin{subfigure}[h]{0.23\textwidth}
        \includegraphics[width=\linewidth]{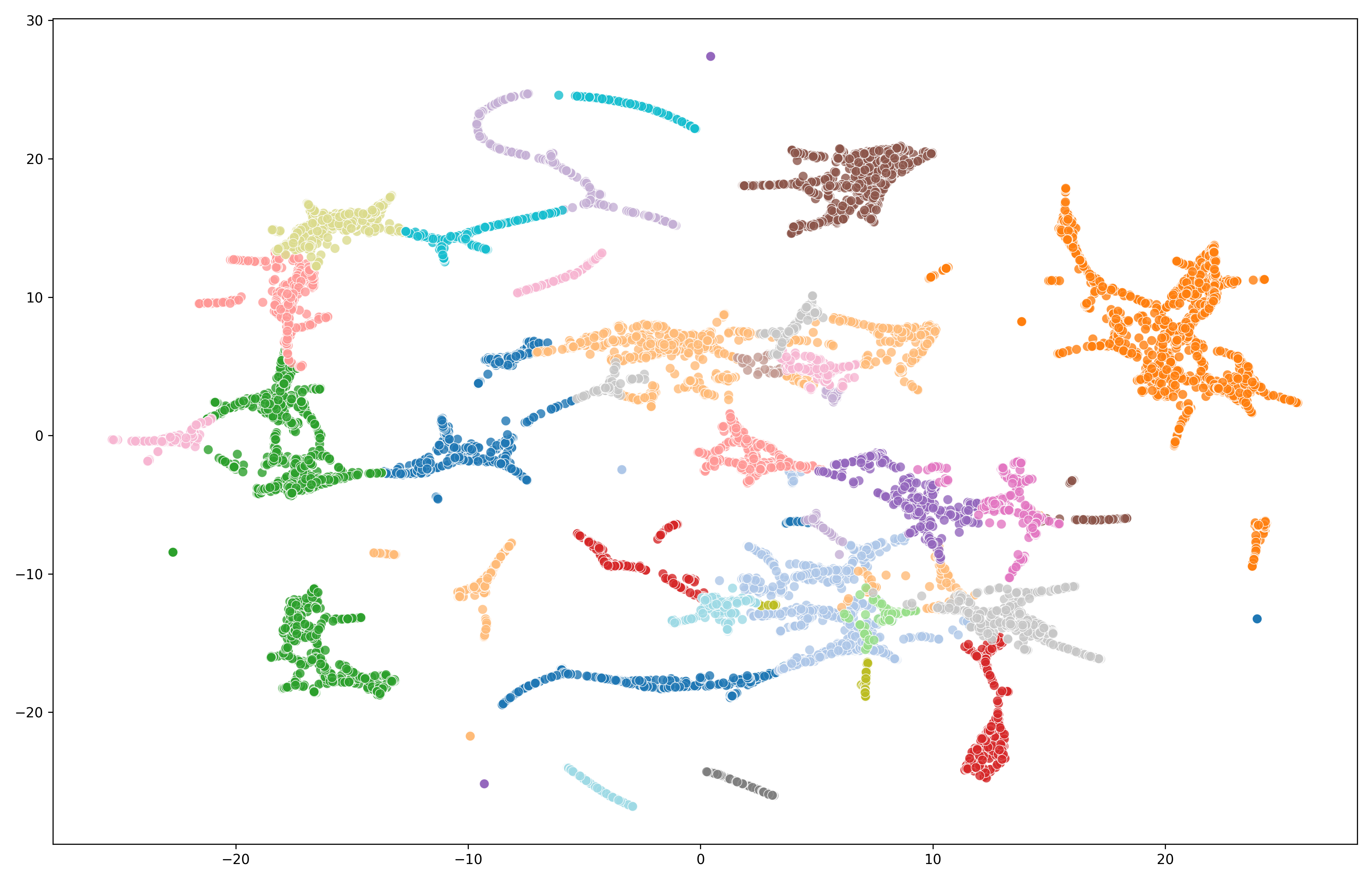}
        \caption{Embeddings for the final cluster assignments.}
        \label{fig:cluster_visuals2}
    \end{subfigure}
    \caption{A visual ablation of the LLM-merging stage on the Bills dataset.}
    \label{fig:cluster_visuals}
\end{figure}

The results in Table \ref{tab:cluster_count} validate our core design principle: beginning with fine-grained, high-purity clusters is important for optimal performance. This is first demonstrated on the Bills dataset, where performance is stable at high initial cluster counts (ARI 0.42-0.43) before dropping at lower granularities. Figure \ref{fig:cluster_visuals} visually depicts this successful consolidation on Bills, contrasting the deliberately fragmented initial assignments (Figure \ref{fig:cluster_visuals1}) with the coherent final topics (Figure \ref{fig:cluster_visuals2}). 
The trend is even more apparent on the Wiki dataset, where starting with 80 clusters (Purity 0.88) yields a final ARI of 0.71, while starting with only 30 (Purity 0.79) causes the ARI to plummet to 0.50.  This confirms that feeding the LLM with coarse, mixed-topic clusters provides ambiguous input, fundamentally compromising the quality of its merge decisions.

However, the data also reveals that resolution, the number of starting clusters, is an equally important factor. A high initial count (e.g., 60-80) provides a fine-grained foundation of small, unambiguous units, allowing the LLM to make a series of simpler, more reliable decisions as it builds topics from the bottom up. In contrast, a low count (e.g., 30-40) forces the LLM to start with large, complex clusters, increasing the risk of making early, high-impact errors, even if those clusters are relatively pure. This effect is particularly evident on the Wiki dataset, where the performance drop between 60 and 40 initial clusters highlights the importance of sufficient granularity. The stable performance observed on both datasets above this threshold further suggests that TopicMerge is robust once a minimum resolution for effective hierarchical reasoning is met.

\section{Conclusion}
The direct application of LLMs to topic modeling has set a new state-of-the-art \cite{wang2023prompting, pham2024topicgpt, liu-etal-2025-llm-guided}, yet these methods remain fundamentally constrained by high computational costs and limited context windows, impeding scalability. In this work, we introduced TopicMerge, a hybrid pipeline that reconciles the global structural awareness of clustering with the local semantic reasoning of an LLM. By delegating corpus organization to an efficient clustering algorithm and reserving the LLM for nuanced semantic arbitration, TopicMerge produces high-quality topics in a scalable manner. Our experiments on the Bills and Wiki benchmarks validate TopicMerge achieves state-of-the-art or highly competitive performance across key metrics (P1, ARI, NMI), while also reducing the number of required LLM prompts by over 92\% compared to document-level models, making high-quality, LLM-enhanced topic modeling practical for real-world datasets.

\section*{Limitations}
While TopicMerge demonstrates a significant advancement in scalable topic modeling, we identify several opportunities for future refinement. The framework's performance is tied to the quality of initial embeddings, presenting a challenge for corpora with very short texts where embeddings can be less discriminative. Our reliance on a closed-source LLM also introduces potential data contamination on benchmark datasets, inherent stochasticity in outputs, and challenges for full reproducibility. Future work could address these limitations by adapting the pipeline for open-source models, which would improve reproducibility and mitigate contamination risks. Furthermore, integrating specialized short-text embedding techniques could expand the method's applicability, making this powerful topic modeling paradigm more accessible. To facilitate further research and help mitigate these reproducibility concerns, our code is made publicly available at \url{https://github.com/pnnl/PermitAI/tree/master/TopicMerge}.

\section*{Ethical Considerations}
While TopicMerge is designed for the general task of topic modeling, its reliance on large language models introduces  the potential for inherited bias from the underlying LLM, which is trained on vast web-scale corpora containing societal biases. These biases can manifest in the generated topic labels and merge decisions, potentially leading to stereotypical or unfair representations of topics related to demographic groups or sensitive social issues. The opaque nature of these models makes it difficult to audit the source of such biases, complicating efforts at mitigation.

\section*{Acknowledgements}
This work was supported by the Office of Policy, U.S. Department of Energy, and Pacific Northwest National Laboratory, which is operated by Battelle Memorial Institute for the U.S. Department of Energy under Contract DE-AC05–76RL01830. This article has been cleared by PNNL for public release as PNNL-SA-228342.

\bibliography{custom}

\newpage
\appendix
\section{Evaluation Metrics}
\label{app:eval_metrics}
Given a set of ground-truth classes \(X = \{x_1, \dots, x_J\}\) and a set of predicted cluster assignments \(Y = \{y_1, \dots, y_K\}\), we evaluate the alignment between \(Y\) and \(X\) using the following external evaluation metrics. Higher values for these metrics indicate better clustering performance.

\subsection{Purity and P1}
Purity is a simple measure of cluster homogeneity. For each predicted cluster, it finds the most frequent ground-truth class and calculates the proportion of documents from that class. The overall purity is the weighted average of these proportions.  We can represent purity as:
\begin{equation}
\text{P}(X, Y) = \frac{1}{N} \sum_{k=1}^{K} \max_{j} |y_k \cap x_j|.
\end{equation}
While simple, purity does not penalize having many small, pure clusters. To address this, we also use the Harmonic Mean of Purity (P1), which balances purity and inverse purity, ensuring clusters are both homogeneous and comprehensive \cite{amigo2009comparison}.  We can represent P1 as:
\begin{equation}
\text{P1} = \sum_{k=1}^{K} \frac{|y_k|}{N} \max_{j} F(y_k, x_j).
\end{equation}
where \(F(y_k, x_j)\) is the F-score for cluster \(y_k\) and class \(x_j\), and \(N\) is the total number of documents:
\begin{equation}
F(y_k, x_j) = \frac{2 \cdot \text{Precision}(y_k, x_j) \cdot \text{Recall}(y_k, x_j)}{\text{Precision}(y_k, x_j) + \text{Recall}(y_k, x_j)}.
\end{equation}

\subsection{Adjusted Rand Index}
The Rand Index (RI) calculates the proportion of pairs of documents that are consistently treated in both the predicted and true clusterings (i.e., either in the same cluster in both, or in different clusters in both). The Adjusted Rand Index (ARI) improves upon RI by correcting for the possibility of chance agreement. It compares the observed RI to the expected RI under a random assignment model. We can represent ARI as:
\begin{equation}
\text{ARI}(X,Y) = \frac{\text{RI} - E[\text{RI}]}{\max(\text{RI}) - E[\text{RI}]}.
\end{equation}
where the Rand Index is defined as:
\begin{equation}
\text{RI}(X,Y) = \frac{\text{TP} + \text{TN}}{\text{TP} + \text{FP} + \text{FN} + \text{TN}}.
\end{equation}
Here, TP (True Positives) is the number of pairs of documents in the same cluster in both \(X\) and \(Y\), and TN (True Negatives) is the number of pairs in different clusters in both. ARI values range from -1 to 1, where 1 indicates perfect agreement.

\subsection{Normalized Mutual Information}
Mutual Information (MI) measures the mutual dependence between the ground-truth labels \(X\) and the predicted clusters \(Y\), quantifying the information shared between them.  We can represent MI as:
\begin{equation}
\text{MI}(X,Y) = \sum_{x \in X} \sum_{y \in Y} p(x, y) \log \left( \frac{p(x, y)}{p(x)p(y)} \right).
\end{equation}
where \(p(x, y)\) is the joint probability distribution, and \(p(x)\) and \(p(y)\) are the marginal probabilities. Normalized Mutual Information (NMI) normalizes this value by the average entropy of both clusterings, to produce a score between 0 and 1. This makes the metric robust to differences in the number of clusters. A score of 1 indicates perfect correlation, while 0 indicates none.  We can represent NMI as:
\begin{equation}
\text{NMI}(X,Y) = \frac{\text{MI}(X,Y)}{\frac{H(Y) + H(X)}{2}}
\end{equation}
where $H(Y)$ and $H(X)$ are the entropy of $X$ and $Y$.

\section{Artifacts, Licensing, and Intended Use}
\label{app:artifacts}
This section details the datasets, models, and software used in this research, their respective licenses, and confirms that their use is consistent with their intended purpose. Our goal is to ensure full transparency and reproducibility.

\subsection{Datasets}
\paragraph{Bills and Wiki.}
The \textit{Bills} and \textit{Wiki} datasets \cite{hoyle2022neural} are standard academic benchmarks designed for evaluating topic modeling and other text categorization tasks. They are publicly available for research purposes, and our use of them for evaluating clustering performance is fully consistent with their intended use. Furthermore, as these are well-established public corpora that have been vetted and used in numerous prior publications at major NLP venues, they are considered to pose minimal risk of containing private, personally identifiable information or significant offensive content

\subsection{Models and Software}
\paragraph{Gemini Models.}
The embedding generation (gemini-embedding-001) and LLM reasoning (Gemini 2.5 Flash Lite and Gemini 3.1 Flash Lite) were performed using models provided by Google. Unless explicitly noted otherwise (as in Table~\ref{tab:results_simple}), all experiments, prompt measurements, and ablations use Gemini 2.5 Flash Lite. These are proprietary models accessed via an API and are subject to Google APIs Terms of Service and Gemini API Additional Terms of Service. Their intended use is for a wide range of natural language processing tasks. Our application of these models for embedding, labeling, and semantic arbitration falls squarely within this intended scope.

\paragraph{PaCMAP.}
The implementation of Pairwise Controlled Manifold Approximation (PaCMAP) is from its official Python package \cite{JMLR:v22:20-1061}, which is distributed under the Apache-2.0 License. Its intended use is for dimensionality reduction, and we use it for exactly this purpose prior to clustering.

\paragraph{Agglomerative Clustering.}
Our initial clustering was performed using the Agglomerative Clustering implementation from the `scikit-learn` library distributed under the BSD-3-Clause License. The intended use of this algorithm is to perform hierarchical clustering, which is consistent with our methodology of generating an initial set of micro-clusters.

\paragraph{DSPy Framework.}
The prompts for LLM interaction were managed using the DSPy framework \cite{khattab2024dspy}. DSPy is an open-source project distributed under the MIT License. Its intended purpose is to provide a structured way to program and optimize LLM-based pipelines, and our use of it to manage our labeling and merging prompts aligns with this goal.

\paragraph{Baseline Models.}
All baseline model results were obtained from \citet{liu-etal-2025-llm-guided}.

\section{Prompts}
\label{app:prompts}
This section provides the exact prompts used for the cluster labeling and merging stages of the TopicMerge pipeline that interfaces with the DSPy framework \cite{khattab2024dspy}.
\begin{table*}[htbp]
\centering
\small
\label{tab:prompt_labeling}
\begin{tabular}{|p{0.95\textwidth}|}
\hline
\textbf{Prompt for Labeling} \\
\hline
You are an expert content analyst and text categorizer.\\\\

Your goal is to propose a concise, descriptive topic label that accurately reflects the **dominant and most central theme** of the cluster. This label should capture the essence of the *majority* of the summaries, allowing for the possibility that a small percentage of summaries might be outliers that do not perfectly align.
**Granularity Target:** The following are examples of the level of specificity your labels should aim for or exceed: `high\_level\_topic\_examples`. Your label should be *at least* as specific as these examples — never broader. These represent the coarsest acceptable level; prefer more specific labels when the texts support it. If the cluster's dominant theme clearly falls under one of these examples, use that example as your label.\\\\

Each label must adhere to the following requirements:\\
1.  **Concise:** Be 1-3 words long.\\
2.  **Optimal Granularity \& Specificity:** Represent the *most specific and relevant conceptual category* that accurately describes the primary topic covered by a majority of the provided summaries. Prioritize specificity for the core theme over attempting to encompass every single outlier. Your label must never be broader than the examples in the granularity target above — if your label would subsume multiple of those examples, it is too broad.\\
3.  **Applicable to Dominant Theme:** The label must accurately and logically apply to the *dominant theme* represented by the majority of summaries in the batch. It is acceptable if a small percentage of outlier summaries do not perfectly fit this label.\\
4.  **Singular Concept:** Each label should describe a single primary conceptual category, not a combination (e.g., "Medieval Architecture" is good; "Places and Institutions" is not).\\
5.  **Subject Matter, Not Format:** The label must describe *what the texts are about*, not how the texts are structured. Labels like "Biographies," "Lists," "Descriptions," "Events," or "Individuals" describe formats or containers — instead, identify the substantive domain (e.g., "American Politicians" instead of "Biographies and Individuals"; "Gothic Cathedrals" instead of "Descriptions of Buildings").\\
6.  **Predictive of Domain:** Ask yourself: "If someone saw only this label, would they correctly predict the single academic field or subject area these texts belong to?" If the label could plausibly point to multiple unrelated fields, it is too vague or misleading.\\\\

Topic Label Description Requirements:\\
1.  **Concise:** Be 1-2 sentences long.\\
2.  **Scope:** Clearly define the scope of the topic label, explaining the specific conceptual area or field of activity it represents. This description should reflect the primary theme identified for the majority of the cluster members.\\
3.  **General \& Conceptual:** Focus on defining the label as a concept. Avoid using specific individuals, document identifiers, names, events, or narrow sub-topics *from the provided summaries* as examples within the description itself.\\
4.  **Tightly Focused:** The description must reinforce the same core domain as the label. Do not mention secondary themes, tangential content, or adjacent fields that only apply to a minority of texts — this creates misleading associations.\\\\

Do not include:\\
1.   Specific individuals, organizations, document identifiers, project names, or specific titles from the provided text *as examples or illustrative details within your description*.\\
2.   Overly obscure specialized jargon or technical terms, unless that jargon *is* the precise conceptual category itself.\\
3.   Secondary or tangential themes that only apply to outlier texts in the cluster.\\
4.   Any conversational remarks or information beyond the requested label and its description.\\

\hline
\end{tabular}
\end{table*}

\begin{table*}[htbp]
\centering
\small
\label{tab:prompt_merging}
\begin{tabular}{|p{0.95\textwidth}|}
\hline
\textbf{Prompt for Merging} \\
\hline
You are deciding whether `Cluster\_A` and `Cluster\_B` should be merged into a single cluster.\\\\

**Goal:** Consolidate fragments of the same topic while keeping genuinely different topics apart.\\\\

**Calibration:**\\
Look at `high\_level\_topic\_examples` — these are real topic labels from this dataset (though not exhaustive). They define the TARGET granularity level your clusters should reach. Your merged clusters should be roughly at this level of breadth. Use `local\_context\_clusters` and `global\_context\_clusters` to understand the broader landscape of topics in this dataset.\\\\

**MERGE if any of these clearly apply:**\\
1. The clusters are about the same core topic or one is a clearly a subset/specialization of the other.\\
2. The clusters are redundant splits of the same concept.\\
3. Both clusters are sub-topics that would naturally fall under a single entry at the granularity level shown by `high\_level\_topic\_examples`.\\
4. Both clusters share the same subject domain AND the same type of content.\\\\

**Do NOT merge if any of these clearly apply:**
1. The clusters would belong to DIFFERENT entries at the granularity level shown by `high\_level\_topic\_examples`.\\
2. The clusters represent fundamentally different real-world domains (e.g., biology vs. economics).\\
3 One cluster is coherent and the other is noise — merging would dilute it.\\\\

**Final Decision:** Only approve a merge if the evidence for it is compelling and unambiguous. If there is any reasonable doubt, keep them separate.\\
\hline
\end{tabular}
\end{table*}

\begin{table*}[htbp]
\centering
\label{tab:prompt_relabeling}
\small
\begin{tabular}{|p{0.95\textwidth}|}
\hline
\textbf{Prompt for Relabeling} \\
\hline
You are an expert content analyst and text categorizer.\\\\

This cluster was formed by merging `cluster\_A` and `cluster\_B`. These prior labels indicate the full scope of content in this cluster. Use them to understand the breadth of what this cluster contains, but do NOT simply concatenate them. Your job is to synthesize a single unified label that naturally encompasses both, guided by the representative `texts` provided. You may generalize, refine, or completely rename if the texts suggest a more accurate characterization.\\\\

**Your Goal:** Propose a concise, descriptive topic label that accurately reflects the **dominant and most central theme** of the combined cluster. This label should capture the essence of the *majority* of the texts, allowing for the possibility that a small percentage might be outliers.\\\\

**Granularity Target:** The `high\_level\_topic\_examples` are real topic labels from this dataset (though not exhaustive). Your label should be *at least* as specific as these examples — never broader. These represent the coarsest acceptable level; prefer more specific labels when the texts support it. If the cluster's dominant theme clearly falls under one of these examples, use that example as your label.\\\\

Each label must adhere to the following requirements:\\
1.  **Concise:** Be 1-3 words long.\\
2.  **Optimal Granularity \& Specificity:** Represent the *most specific and relevant conceptual category* that accurately describes the primary topic covered by a majority of the provided texts. Your label must never be broader than the examples in `high\_level\_topic\_examples` — if your label would subsume multiple of those examples, it is too broad.\\
3.  **Unified Coverage:** The label must naturally encompass the content from both `cluster\_A` and `cluster\_B`. Do not pick a label that only covers one source cluster while ignoring the other.\\
4.  **Singular Concept:** Each label should describe a single primary conceptual category, not a combination (e.g., "Medieval Architecture" is good; "Places and Institutions" is not).\\
5.  **Subject Matter, Not Format:** The label must describe *what the texts are about*, not how the texts are structured. Labels like "Biographies," "Lists," "Descriptions," "Events," or "Individuals" describe formats or containers — instead, identify the substantive domain (e.g., "American Politicians" instead of "Biographies"; "Gothic Cathedrals" instead of "Descriptions of Buildings").\\
6.  **Predictive of Domain:** Ask yourself: "If someone saw only this label, would they correctly predict the single academic field or subject area these texts belong to?" If the label could plausibly point to multiple unrelated fields, it is too vague.\\
7.  **Informed Synthesis:** Use `cluster\_A` and `cluster\_B` labels to understand scope, but let the representative `texts` ground your final decision. If the texts reveal that both source clusters are actually about the same specific thing, prefer that specific label over a forced generalization.\\\\

Topic Label Description Requirements:\\
1.  **Concise:** Be 1-2 sentences long.\\
2.  **Scope:** Clearly define the scope of the topic label, explaining how the content from both source clusters coheres under this single label.
3.  **General \& Conceptual:** Focus on defining the label as a concept. Avoid using specific individuals, document identifiers, names, events, or narrow sub-topics from the provided texts as examples.\\
4.  **Tightly Focused:** The description must reinforce the same core domain as the label. Do not mention secondary themes or tangential content that only applies to a minority of texts.\\\\

Do not include:\\
1.   Specific individuals, organizations, document identifiers, project names, or specific titles from the provided texts.\\
2.   Overly obscure specialized jargon, unless that jargon *is* the precise conceptual category itself.\\
3.   Secondary or tangential themes that only apply to outlier texts.\\
4.   Any conversational remarks or information beyond the requested label and its description.\\
\hline
\end{tabular}
\end{table*}

\newpage
\section{Topic Labels}
\label{app:topics}
We provide the topic labels generated by TopicMerge using our default Gemini 2.5 Flash Lite backbone in Table~\ref{tab:topic}. A key observation is that the model's output is more granular than the high-level ground-truth categories for both the Bills and Wiki datasets. This outcome reflects the LLM's tendency to prioritize fine-grained semantic purity during the merging process. This presents an opportunity for future research to develop methods for controlling topic abstraction, for instance by using few-shot examples to guide the LLM or by exploring different merging heuristics to better align with predefined topic scopes.

\begin{table*}[htbp]
\centering
\small
\caption{The topic labels for the Bills and Wiki datasets.}
\setlength{\tabcolsep}{3pt}
\begin{tabular}{l|ll}
\hline\hline
\textbf{Data} & \multicolumn{2}{c}{\textbf{Topic Labels}} \\
\hline
\multirow{29}{*}{Bills}
& \multicolumn{2}{c}{Ground Truth} \\
\cline{2-3}
& health: 2316              & government operations: 1937  \\
& domestic commerce: 1323   & Defense: 1297 \\
& Public Lands: 1232        & Law and Crime: 967 \\
& Environment: 692          & Transportation: 663 \\
& Energy: 661               & Macroeconomics: 610 \\
& International Affairs: 554 & Labor: 493 \\
& Foreign Trade: 471        & Education: 448 \\
& Social Welfare: 436       & Civil Rights: 353 \\
& Technology: 321           & Housing: 306 \\
& Agriculture: 162          & \\
\cline{2-3}
& \multicolumn{2}{c}{TopicMerge} \\
\cline{2-3}
& Disaster Response and Preparedness: 530 & Government Operations \& Reform: 498 \\
& Housing and Family Support Policy: 245 & Federal Economic and Financial Legislation: 1332 \\
& Energy Policy and Infrastructure: 774 & Environmental Protection and Management: 878 \\
& Healthcare and Public Health Legislation: 2124 & Fiscal Policy and Deficit Reduction: 414 \\
& Education Policy: 391 & Immigration and Security Legislation: 667 \\
& Workforce Development and Training: 182 & Veterans Affairs Legislation: 929 \\
& Government Funding and Defense Acquisition: 377 & Space Exploration \& Aeronautics: 509 \\
& International Trade and Economic Policy: 318 & Cybersecurity and Information Technology: 105 \\
& Small Business Innovation and Technology: 101 & Historical Commemorations and Accountability: 256 \\
& Crime and Trafficking Legislation: 490 & Federal Labor and Employment: 321 \\
& Foreign Policy \& State Department: 272 & Law Enforcement and Public Safety: 411 \\
& Currency and Coins: 208 & Voting Rights and Election Administration: 143 \\
& Social Security Benefits: 62 & Official English Laws: 20 \\
& Agriculture and Food Policy: 448 & FCC Regulations: 103 \\
& Wage and Hour Laws: 117 & Historic Sites and Parks: 582 \\
& Transportation Safety: 688 & Employment Discrimination: 70 \\
& Public Lands \& Indigenous Affairs: 518 & Intellectual Property Law: 159 \\
\hline
\multirow{28}{*}{Wiki}
& \multicolumn{2}{c}{Ground Truth} \\
\cline{2-3}
& Media and drama: 1217         & Sports and recreation: 1018 \\
& Warfare: 978                  & Music: 976 \\
& Natural sciences: 952         & Engineering and technology: 615 \\
& Social sciences and society: 589 & History: 446 \\
& Video games: 362              & Geography and places: 276 \\
& Language and literature: 235   & Art and architecture: 199 \\
& Philosophy and religion: 114  & Agriculture, food, and drink: 38 \\
& Mathematics: 9                & \\
\cline{2-3}
& \multicolumn{2}{c}{TopicMerge} \\
\cline{2-3}
& US Cities and Institutions: 66 & Notable individuals: 77 \\
& Military and Scientific History: 1132 & Video Games: 426 \\
& Cities and Towns: 52 & Restaurants and Eateries: 149 \\
& Historical Locations and Structures: 372 & Archaeological Sites: 53 \\
& Government and Politics: 80 & Mythology and Folklore: 108 \\
& US Law and Politics: 170 & Economics and Finance: 17 \\
& Sports: 1008 & Fictional Works and Media: 1398 \\
& Educational Institutions: 118 & Science and Psychology: 120 \\
& English Law Cases: 62 & Musicians: 981 \\
& Urban settlements: 74 & Scottish History and Culture: 68 \\
& Geology and Earth Science: 5 & Norse Mythology Figures: 45 \\
& Biology and Medicine: 171 & Chemistry Elements and Reactions: 86 \\
& Aviation Incidents and Infrastructure: 123 & Computer Software and Operating Systems: 38 \\
& European Royalty and Nobility: 38 & Dinosaurs: 32 \\
& Hindu Mythology Figures: 43 & Billiards and Pool: 12 \\
& Weather events: 269 & Mammals: 121 \\
& Astronomy and Space Science: 145 & Highways and roads: 357 \\
& Historic Infrastructure and Vessels: 8 & \\
\hline\hline
\end{tabular}%
\label{tab:topic}
\end{table*}

\section{Exploring Greedy Ground-Truth Guided Merging with TopicMerge}
To establish a practical upper bound for our merging algorithm, we conducted an oracle experiment using the same hyperparameters from Section \ref{sec:experimental_setup}. In this setup, we replaced the LLM arbiter with a ground-truth guided heuristic that greedily merged any two clusters sharing the same majority ground-truth class, effectively simulating a near-perfect semantic arbiter. The results, presented in Table \ref{tab:results_optimal}, demonstrate a substantial improvement over all other methods. This significant performance gap provides compelling evidence that the core pipeline of TopicMerge is fundamentally sound and has a very high performance ceiling. The primary limiting factor is the LLM's ability to perfectly replicate the oracle's optimal merge decisions. Furthermore these findings are not a theoretical maximum for two key reasons. First, the merging heuristic is greedy, meaning a more optimal, non-greedy algorithm could yield even better results. Second, this experiment does not explore further refinements to the initial cluster creation stage where improvements in embeddings, dimensionality reduction, and clustering would provide an even stronger starting point for the merging process. This indicates that the true potential of our hybrid pipeline is even greater, representing an exciting direction for future research.

\begin{table*}[ht]
\centering
\small
\caption{Experimental results comparing TopicMerge with greedy ground-truth guided merging against baselines. The best results among standard methods are marked in \textbf{bold} and second-best are \underline{underlined}; oracle results are shown for upper-bound comparison.}
\label{tab:results_optimal}
\begin{tabular}{@{}l ccc ccc@{}}
\toprule
& \multicolumn{3}{c}{\textbf{Bills}} & \multicolumn{3}{c}{\textbf{Wiki}} \\
\cmidrule(lr){2-4} \cmidrule(lr){5-7}
& \textbf{P1} & \textbf{ARI} & \textbf{NMI} & \textbf{P1} & \textbf{ARI} & \textbf{NMI} \\
\midrule
LDA & 0.52 & 0.31 & 0.42 & 0.72 & 0.63 & 0.71 \\
NMF & 0.24 & 0.04 & 0.15 & 0.58 & 0.38 & 0.55 \\
CTM & 0.37 & 0.19 & 0.32 & 0.61 & 0.46 & 0.62 \\
FASTopic & 0.40 & 0.30 & 0.33 & 0.65 & 0.56 & 0.60 \\
EDTM & \textbf{0.58} & — & — & 0.65 & — & — \\
BERTopic & 0.41 & 0.18 & 0.37 & 0.62 & 0.50 & 0.59 \\
PromptTopic & 0.36 & 0.17 & 0.28 & 0.62 & 0.46 & 0.59 \\
TopicGPT  & \underline{0.57} & \underline{0.40} & 0.49 & \textbf{0.74} & 0.60 & 0.70 \\
LiSA & \underline{0.57} & \underline{0.40} & 0.51 & \textbf{0.74} & \underline{0.66} & 0.71 \\
\midrule
\multicolumn{7}{@{}l}{\textbf{TopicMerge (Ours)}} \\
\quad \textit{w/ 2.5 Flash-Lite} & \underline{0.57} & \textbf{0.42} & \textbf{0.53} & 0.72 & \textbf{0.71} & \underline{0.73}\\
\quad \textit{w/ 3.1 Flash-Lite} & \underline{0.57} & \textbf{0.42} & \underline{0.52} & \underline{0.73} & \textbf{0.71} & \textbf{0.74}\\
\midrule
\multicolumn{7}{@{}l}{\textbf{Oracle Upper Bound}} \\
\quad \textit{TopicMerge (ground-truth)} & \textit{0.66} & \textit{0.49} & \textit{0.56} & \textit{0.87} & \textit{0.81} & \textit{0.79}\\
\bottomrule
\end{tabular}
\end{table*}

\end{document}